\documentclass{article}
\usepackage{arxiv}
\usepackage[utf8]{inputenc} 
\usepackage[T1]{fontenc}    
\usepackage{url}  
\usepackage{hyperref}       
\hypersetup{
    colorlinks=true,       
    linkcolor=blue,          
    citecolor=blue,        
    urlcolor=blue           
}
\usepackage{booktabs}       
\usepackage{nicefrac}       
\usepackage{microtype}      
\usepackage{lipsum}		
\usepackage{graphicx}
\usepackage{doi}
\usepackage{amsmath}
\usepackage{subfig}
\usepackage[affil-it]{authblk}
\usepackage[english]{babel}
\usepackage{blindtext}
\usepackage{amssymb}
\usepackage{cleveref}
\usepackage{booktabs}
\usepackage{framed,multirow}
\usepackage{hyperref}
\usepackage{url}            
\usepackage{booktabs}
\usepackage[mathlines]{lineno}
\usepackage{amsfonts}       
\usepackage{latexsym}
\usepackage{amsmath}
\usepackage{amssymb}
\usepackage{amsfonts}

\usepackage{cleveref}
\usepackage{nicefrac}       
\usepackage{microtype}      
\usepackage{lipsum}
\usepackage{graphicx}
\usepackage{adjustbox}
\usepackage{graphicx}
\usepackage[table,xcdraw]{xcolor}
\definecolor{newcolor}{rgb}{.8,.349,.1}
\usepackage[linesnumbered,ruled,vlined]{algorithm2e}

\usepackage{multirow}
\usepackage{diagbox}
\usepackage{subfig}
\usepackage{tikz}
\usetikzlibrary{external}

\usepackage{pgfplots}
\pgfplotsset{compat=1.7}
\usetikzlibrary{positioning}
\usepackage{etoolbox} 
\usepackage[numbers]{natbib}

\newcommand*\linenomathpatch[1]{%
  \cspreto{#1}{\linenomath}%
  \cspreto{#1*}{\linenomath}%
  \csappto{end#1}{\endlinenomath}%
  \csappto{end#1*}{\endlinenomath}%
}
\newcommand*\linenomathpatchAMS[1]{%
  \cspreto{#1}{\linenomathAMS}%
  \cspreto{#1*}{\linenomathAMS}%
  \csappto{end#1}{\endlinenomath}%
  \csappto{end#1*}{\endlinenomath}%
}
\expandafter\ifx\linenomath\linenomathWithnumbers
  \let\linenomathAMS\linenomathWithnumbers
  \patchcmd\linenomathAMS{\advance\postdisplaypenalty\linenopenalty}{}{}{}
\else
  \let\linenomathAMS\linenomathNonumbers
\fi

\linenomathpatch{equation}
\linenomathpatchAMS{gather}
\linenomathpatchAMS{multline}
\linenomathpatchAMS{align}
\linenomathpatchAMS{alignat}
\linenomathpatchAMS{flalign}

\SetCommentSty{mycommfont}

\crefformat{section}{\S#2#1#3}
\crefformat{subsection}{\S#2#1#3}
\crefformat{subsubsection}{\S#2#1#3}
\crefrangeformat{section}{\S\S#3#1#4 to~#5#2#6}
\crefmultiformat{section}{\S\S#2#1#3}{ and~#2#1#3}{, #2#1#3}{ and~#2#1#3}

\date{}
\title{A Generalized Schwarz-type Non-overlapping Domain Decomposition Method using Physics-constrained Neural Networks}

\author{\href{https://orcid.org/0000-0002-1095-0881}{\includegraphics[scale=0.08]{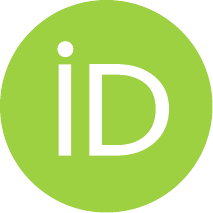}\hspace{1mm}Shamsulhaq Basir}\thanks{~shb105@pitt.edu (Shamsulhaq Basir)}}

\author{\href{https://orcid.org/0000-0003-1967-7583}{\includegraphics[scale=0.08]{orcid.pdf}\hspace{1mm}Inanc Senocak\thanks{corresponding author:~senocak@pitt.edu (Inanc Senocak)}}}

\affil{Department of Mechanical Engineering and Materials Science, University of Pittsburgh, \\ 3700 O'Hara St., Pittsburgh, PA 15261, USA}
\renewcommand{\shorttitle}{\title{A Generalized Schwarz-type Non-overlapping DDM using PECANNs}}

\begin{document}
\maketitle
\begin{abstract}
We present a meshless Schwarz-type non-overlapping domain decomposition method based on artificial neural networks for solving forward and inverse problems involving partial differential equations (PDEs). To ensure the consistency of solutions across neighboring subdomains, we adopt a generalized Robin-type interface condition, assigning unique Robin parameters to each subdomain. These subdomain-specific Robin parameters are learned to minimize the mismatch on the Robin interface condition, facilitating efficient information exchange during training. Our method is applicable to both the Laplace's and Helmholtz equations. It represents local solutions by an independent neural network model which is trained to minimize the loss on the governing PDE while strictly enforcing boundary and interface conditions through an augmented Lagrangian formalism. A key strength of our method lies in its ability to learn a Robin parameter for each subdomain, thereby enhancing information exchange with its neighboring subdomains. We observe that the learned Robin parameters adapt to the local behavior of the solution, domain partitioning and subdomain location relative to the overall domain. Extensive experiments on forward and inverse problems, including one-way and two-way decompositions with crosspoints, demonstrate the versatility and performance of our proposed approach.
\end{abstract}



\section{Introduction}
Deep learning with artificial neural networks (ANNs) has transformed many fields of science and engineering. The functional expressivity of ANNs was established by universal approximation theory. Since then ANNs have emerged as a meshless method to solve partial differential equations (PDEs) for both forward and inverse problems \cite{dissanayake1994neural, van1995neural, Monterola1998lagrange, lagaris1998artificial}. With the introduction of easily accessible software tools for auto-differentiation and optimization, the use of ANNs to solve PDEs has grown rapidly in recent years as physics-informed neural networks (PINNs) \cite{raissi2017physicsII, Karniadakis2021}. Numerous works have been published since the introduction of the PINN framework to address the shortcomings of the framework as well as expand it with different features such as uncertainty quantification. 

PINNs offer several advantages over conventional numerical methods such as the finite element and volume methods when applied to data-driven modeling, and inverse and parameter estimation problems. Unlike conventional numerical methods that have been developed and advanced over several decades as predictive-science techniques for challenging problems, PINNs have thus far been mostly applied to two-dimensional problems. Several issues stand in the way of extending PINNs to large, three-dimensional, multi-physics problems, including difficulties with nonlinear non-convex optimization, respecting conservation laws strictly, and long training times. In the present work, we focus on the application of domain decomposition methods (DDM) to PINNs, which are motivated by solving forward and inverse problems that can be computationally large and may involve multiple physics.  

\begin{figure}
    \centering
\includegraphics[scale=0.6]{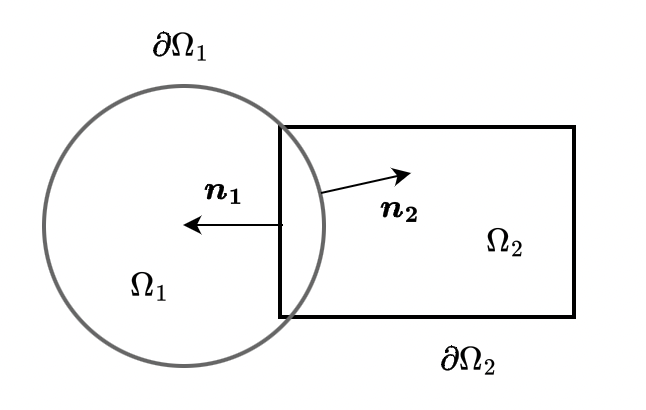}\quad
\includegraphics[scale=0.6]{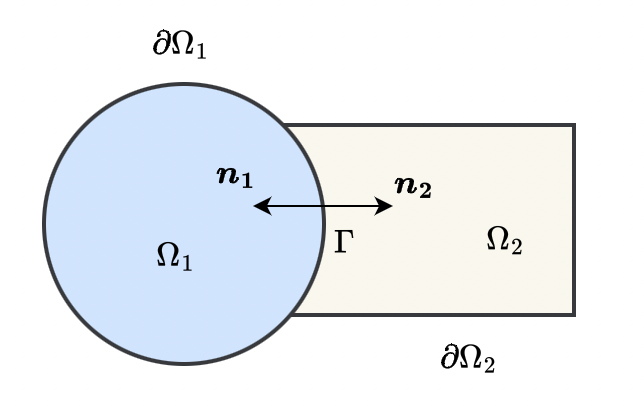}
    \caption[domain decomposition types]{domain decomposition types: (a) overlapping subdomains, (b) non-overlapping subdomains}
    \label{fig:overlapping_non_overlapping_domains}
\end{figure}

Domain decomposition has become an essential strategy for solving complex PDE problems that are too large to be solved on a single computer or that have complex geometries with multiple physics \cite{dolean2015introduction}. Domain decomposition methods can be constructed as overlapping or non-overlapping as shown in Fig. \ref{fig:overlapping_non_overlapping_domains}a and Fig. \ref{fig:overlapping_non_overlapping_domains}b, respectively. There are different type of domain decomposition techniques. A detailed discussion of these methods can be found in textbooks written on the subject matter \cite{quarteroni1999DDM, smith1998domain, dolean2015introduction}. In the present study, we will focus on Schwarz-type methods, specifically optimized Schwarz methods \cite{japhet1998, gander2006optimized}. 
Most of the modern developments in DDM have taken place with conventional numerical methods such as finite element and volume methods in mind. On the other hand, domain decomposition in the context of PINNs is a new and active research area that has been the subject of several recent works. 
\citet{li2019d3m} proposed an overlapping domain decomposition method based for the DeepRitz method \cite{weinan2017proposal}  which is an alternative formulation of PINNs for learning the solution of PDEs. In their approach, the alternating Schwarz method with a Dirichlet-type overlapping interface condition was used and the arising loss term was incorporated into the objective function of the DeepRitz method. Similar to the work of \cite{li2019d3m},  \citet{li2020ddmelliptic} solved Poisson's equation on overlapping decomposed domains with a complex interface using the baseline PINN approach. In their approach a classical alternating Schwarz type method was used as well and the loss term arising from satisfying the interface conditions was added to the PINN's objective function in a composite fashion along with loss terms arising from the residual forms of the boundary conditions and the PDE. These works \cite{li2019d3m,li2020ddmelliptic} demonstrated the feasibility of using Schwarz type domain decomposition methods in the context of PINNs.
Recently, \citet{dolean2022finite} introduced finite bases physics-informed neural networks (PINNs) for solving PDEs on overlapping subdomains. Specifically, the authors pursued a Schwarz-type domain decomposition approach in which a PINN model is trained for each subdomain. However, to improve the accuracy of the local solutions, the authors also trained a neural network for the entire domain to serve as a coarse correction. 

\citet{jagtap2020conservative} decomposed a spatial domain into smaller domains and used the baseline PINN method to learn the solution of a PDEs on the whole domain. A separate neural network was adopted in each subdomain and flux continuity across subdomain interfaces were enforced in strong form. The average value of the solution between two subdomains sharing an interface was also enforced as an additional condition. Since, the neural network models associated for subdomains exchange information at each epoch, makes it not strictly Schwarz-type domain decomposition method. In the spirit of the baseline PINNs, loss terms arising from the flux continuity across the subdmoain interfaces are lumped into a single composite objective function with tunable weights. In a followup work, \citet{jagtap2020xpinns} extended the work presented in \cite{jagtap2020conservative} to include the time domain. Furthermore, in this followup work, the interface conditions were simplified to make the method applicable to PDEs that may not represent conservation laws. A parallel implementation of these works is presented in \cite{shukla2021parallel_pinns} showing decent scalability and speedup.

Clearly, domain decomposition in the context of scientific machine learning or physics-informed neural networks is a growing area of focus. Success in this front is expected to enable neural networks to tackle larger problems or reduce training times substantially. Additionally, empirical evidence shows that training separate neural networks on smaller domains is much more feasible and likely to converge than training a single neural network on a large domain with many points. In what follows, we present the theory behind the optimized Schwarz methods \cite{gander2006optimized} and our physics- and equality-constrained artificial networks (PECANN) for solving forward and inverse PDE problems \cite{basir2023adaptive}. We then propose a non-overlapping generalized Schwarz-type domain decomposition method with a Robin-type interface condition with learnable, subdomain-specific parameters for our PECANN framework. We then apply the resulting method to learn the solution of forward and inverse PDE problems using decomposed domains with increasing complexity.  

\section{Technical Background}
The earliest example of an overlapping domain decomposition method is the work of \citet{schwarz1870uber}, which later became known as the alternating Schwarz method (ASM). 
Multiplicative Schwarz method (MSM) is a generalization of ASM. It involves solving a PDE on the first subdomain and then on the second subdomain sequentially. Values from neighboring subdomains at the most current iteration are used as interface conditions during each solve. A drawback of MSM is that it is not amenable to parallelization because of its sequential nature. Additive Schwarz method \cite{additive_schwarz} is a slight modification of the MSM that enables parallel computations by solving the problem on both subdomains concurrently using the information from the previous iteration \cite{quarteroni1999DDM}. However, these aforementioned variants of the Schwarz method are computationally slow and do not converge for non-overlapping subdomains \cite{dolean2002optimized}. Furthermore, these methods do not converge for acoustic problems despite overlapping subdomains \cite{gander2006optimized}. \citet{lions1988schwarz} proposed replacing the Dirichlet boundary conditions in ASM for Laplace's equations with Robin boundary conditions, whereas for Helmholtz equation, \citet{despres1990} proposed radiation conditions. These modifications extended ASM to non-overlapping domains while being applicable to overlapping domains. \citet{japhet1998} optimized the transmission conditions to achieve faster convergence, which has become known as the optimized Schwarz method (OSM) \cite{gander2006optimized}. Although optimal conditions leading to the best convergence in OSM are tied to the Steklov–Poincaré operator \cite{Chevalier1998SymmetrizedMW}, their non-local nature makes them challenging to implement efficiently in numerical simulations \cite{MadayOSM2007}. Hence, the alternative approach involves approximating optimal conditions using local operators, which can then be fine-tuned for enhancing the convergence of OSM \cite{lions1988schwarz}.

There are other domain decomposition methods besides the Schwarz-type methods. For instance, substructuring algorithms such as balancing domain decomposition (BDD) methods \cite{mandel1993balancing} and finite element tearing and interconnect (FETI) \cite{farhat1991method} are domain decomposition methods to solve a system of linear equations arising from the finite element discretization. For BDD, FETI, and other type of domain decomposition methods, we refer the reader to textbooks dedicated to the subject matter \cite{smith1998domain, quarteroni1999DDM, dolean2015introduction}. \citet{heinlein2021combining} provides a review of recent FETI-type domain decomposition methods in the context of machine learning.
\subsection{Optimized Schwarz method}\label{sec:osm}

Our proposed domain decomposition method has important parallels with the optimized Schwarz methods, but also differ from the OSM in unique ways. Therefore, we briefly explain OSM and discuss some of the key works in OSM.  

Let us consider a typical second-order elliptic PDE on two subdomains for demonstration purposes. On the first subdomain, we consider
\begin{equation}
\begin{aligned}
    -\nabla (u_1 ^{n+1})  &= s_1 &&\text{in} \quad \Omega_1,\\
    u_1^{n+1} &= 0  &&\text{on} \quad \partial \Omega_1 \cap \partial \Omega,\\
    (\mathcal{A}_1 + \beta_1 \frac{\partial}{\partial \boldsymbol{n_1}}) u_1 ^{n+1} &= (\mathcal{A}_1 + \beta_1 \frac{\partial}{\partial \boldsymbol{n_1}}) u_2 ^{n} &&\text{on} \quad \Gamma_1,
\end{aligned}
\end{equation}
and on the second subdomain
\begin{equation}
\begin{aligned}
    -\nabla (u_2 ^{n+1})  &= s_2 &&\text{in} \quad \Omega_2,\\
    u_2^{n+1} &= 0  &&\text{on} \quad \partial \Omega_2 \cap \partial \Omega,\\
    (\mathcal{A}_2 + \beta_2 \frac{\partial}{\partial \boldsymbol{n_2}}) u_2^{n+1} &= (\mathcal{A}_2 + \beta_2 \frac{\partial}{\partial \boldsymbol{n_2}}) u_1^{n} &&\text{on} \quad \Gamma_2,
\end{aligned}
\end{equation}
where $\boldsymbol{n_1}$ and $\boldsymbol{n_2}$ are the outward normal directions on the subdomain boundaries of $\Omega_1$ and $\Omega_2$, respectively. $\Gamma_1$ and $\Gamma_2$ represents the subdomain interfaces corresponding to $\Omega_1$ and $\Omega_2$, respectively. In the case of a non-overlapping domain decomposition $\Gamma_1$ and $\Gamma_2$ are identical.  $\mathcal{A}_1$ and $\mathcal{A}_2$ are operators that act along the interfaces $\Gamma_1$ and $\Gamma_2$, respectively. $\beta_1$ and $\beta_2$ are real valued functions. With $\beta_1 = \beta_2 = 0$ and $\mathcal{A}_1$ and $\mathcal{A}_2$ being identity operators, the original Schwarz method is recovered. As a remedy to the drawbacks of classical Schwarz methods (MSM, ASM), \citet{lions1990schwarz} proposed to replace Dirichlet interface conditions with Robin interface conditions with a tunable parameter $\alpha$. In the above interface formulation, we see that with $\beta_1 = \beta_2 = 1$ and $\mathcal{A}_1 = \mathcal{A}_2 = \alpha$, where $\alpha>0.0$, we recover the Robin interface conditions proposed by \citet{lions1990schwarz}.

The essence of optimized Schwarz method (OSM) \cite{gander2006optimized} is to determine optimal operators $\mathcal{A}$ and the parameter $\beta$ such that the convergence rate of the Schwarz algorithm is minimized. This is often achieved by theoretically deriving an expression for the convergence rate for a representative problem with a simple decomposition (e.g. two subdomains) and optimizing the interface parameters with respect to that convergence rate. The extension of this approach to complex domains with challenging decompositions with many subdomains is admittedly a formidable task. However, numerical experiments have shown that optimal interface conditions, once derived from canonical problems, can be used in complex problems with a general decomposition, as shown in several works \cite{gander2002optimized, gander2007_two_sided_osm, nataf2005, MadayOSM2007}.  We should note that the parameters (i.e. $\beta$ used in the Robin transmission conditions do not have to be the same for each subdomain. For instance, \citet{gander2007_two_sided_osm} proposed a two-sided Robin condition for the Helmholtz equation on non-overlapping domains in which different parameters were adopted in the Robin transmission conditions adopted in each subdomain. \citet{gander2007_two_sided_osm} attained better convergence rates with two-sided Robin condition compared to using the same parameters in the Robin transmission condition.

\subsection{Physics and Equality Constrained Artificial Neural Networks}
In this section, we present our recently developed physics and equality constrained artificial neural networks (PECANN) as a meshless neural network based solver for forward and inverse PDE problems \cite{PECANN_2022}. We will then introduce a generalized Schwarz-type domain decomposition method with a Robin interface condition and extend our PECANN framework to solve forward and inverse PDE problems with domain decomposition to enable distributed learning.

Let us consider a general constrained optimization problem with equality constraints
\begin{equation}
    \min_{\theta} \mathcal{J}(\theta), ~\quad \text{such that } ~\quad \mathcal{C}_i(\theta) =0, \quad \forall i \in \mathcal{E}, \label{eq:constrained_problem}
\end{equation}
where the objective function $\mathcal{J}$ and the constraint functions $\mathcal{C}_i$ are all smooth, real valued functions on a subset of $R^n$ and $\mathcal{E}$ is a finite set of equality constraints. We can cast the constrained optimization problem \eqref{eq:constrained_problem} into an unconstrained optimization problem using the augmented Lagrangian formalism \cite{hestenes1969multiplier, powell1969method} as follows:
\begin{equation}
    \max_{\lambda} \min_{\theta} \mathcal{L}(\theta,\lambda;\mu) =  \mathcal{J}(\theta) + \sum_{i \in \mathcal{E}} \lambda_i \mathcal{C}_i(\theta)  + \frac{1}{2} \sum_{i \in \mathcal{E}} \mu_i  \mathcal{C}^2_i(\theta), \label{eq:unconst_problem}
\end{equation}
where $\lambda_{i}$ is a vector of Lagrange multipliers and $\mu_{i}$ is a vector of penalty parameters. The minimization of Eq. \ref{eq:unconst_problem} can be performed using a variant of gradient descent type optimizer for a sequence of Lagrange multipliers generated by the following adaptive update strategy proposed in \citet{basir2023adaptive} 
\begin{linenomath}
\begin{align}
    \bar{v}_i & \xleftarrow{} \alpha \bar{v}_i + (1 - \alpha) 
    \mathcal{C}_i(\theta)^2, &&\forall i \in \mathcal{E},\\
    \mu_i & \xleftarrow{} \frac{\gamma}{\sqrt{\bar{v}_i} + \epsilon}, &&\forall i \in \mathcal{E},\\
    \lambda_i &\xleftarrow{} \lambda_i +  \mu_i \mathcal{C}_i(\theta), &&\forall i \in \mathcal{E},
    \label{eq:adaptive_dual_update}
\end{align}
\end{linenomath}
where $\bar{v}_i$ are the weighted moving average of the squared gradient of our Lagrange multipliers, $\gamma$ is a scheduled global learning rate, $\epsilon$ is a term added to the denominator to avoid division by zero for numerical stability and $\alpha$ is a smoothing constant. In Algorithm \ref{alg:adaptive_training_algorithm}, we present the training procedure for PECANNs on a physical domain.
\IncMargin{1em}
\begin{algorithm}[!h]
\SetAlgoLined
\SetKw{KwInput}{Input:}
\SetKw{KwOutput}{Output:}
\SetKw{KwDefaults}{Defaults:}
\KwDefaults{$\gamma = 1\times 10^{-2}, ~\alpha = 0.99,~ \epsilon = 1\times 10^{-8}$}\\

\KwInput{$\theta^0$}\\
$\lambda_{i}^0 = 1 \quad \forall i \in \mathcal{E}$ \hspace{6em}
\tcc{Initializing Lagrange multipliers}
$\mu_{i}^0 = 1  \quad \forall i \in \mathcal{E}$ \hspace{5em}
\tcc{Initializing penalty parameters}
$\bar{v}_{i}^0 = 0 \quad \forall i \in \mathcal{E}$ \hspace{6em}
\tcc{initializing averaged square-gradients}
\BlankLine
\For{$t = 1  ~ \KwTo ...$}{
    $\theta^t \leftarrow \underset{\theta}{\mathrm{argmin}}~ \mathcal{L}(\theta^{t-1};\lambda^{t-1},\mu^{t-1})$\hspace{6em}
    \tcc{primal update}
    $\bar{v}_i^t  \xleftarrow{} \alpha ~\bar{v}_i^{t-1} + (1 - \alpha)~
    \mathcal{C}_i(\theta^t)^2, \quad \forall i \in \mathcal{E}$\hspace{3em}
    \tcc{square-gradient update}
    $\mu_i^{t}  \xleftarrow{} \frac{\gamma}{\sqrt{\bar{v}_i^t} + \epsilon}, \quad \forall i \in \mathcal{E}$\hspace{10.5em}
   \tcc{penalty update}
    $\lambda_i^t \xleftarrow{} \lambda_i^{t-1} +  \mu_i^{t} ~ \mathcal{C}_i(\theta^{t}), \quad \forall i \in \mathcal{E}$ \hspace{6em}
    \tcc{dual update}
    }
\KwOutput{$\theta^t$}\\
 \caption{Adaptive Augmented Lagrangian Method}
 \label{alg:adaptive_training_algorithm}
\end{algorithm}
The input to the algorithm is an initialized set of parameters (i.e, $\theta^{0}$) associated with the neural network model representing the solution on the physical domain, a global learning rate $\gamma$, and a smoothing constant $\alpha$. In Algorithm \ref{alg:adaptive_training_algorithm}, the Lagrange multiplier vector is initialized to $1.0$ with their respective averaged squared-gradients initialized to zero.

We have chosen to employ our PECANN framework due to its inherent strength in formulating and solving forward/inverse PDE problems with given constraints. PECANNs excel in this regard by formulating a constrained optimization problem based on a given PDE, and then utilizing an adaptive augmented Lagrangian method to create an equivalent dual unconstrained optimization formulation that is suitable for neural networks. This unique approach enables PECANNs to effectively address learning problems with constraints. Unlike other methods that rely on heuristics to balance the interplay between objective functions \cite{mcclenny2020self,wang2020understanding}, PECANNs provide a more robust and principled approach. By leveraging the augmented Lagrangian formulation, PECANNs offer a general and systematic approach for incorporating constraints into the learning process, enhancing the overall effectiveness and reliability of the method. 

\section{Proposed Domain Decomposition Method}
In this section, we aim to develop a generalized Schwarz-type domain decomposition method that facilitates distributed learning of both forward and inverse PDE problems using artificial neural networks. To achieve this, we adopt our PECANN framework as a solver for each subdomain. Notably, we consider a generalized Robin-type interface transmission conditions as an additional constraint on the solution of each subdomain. By incorporating these transmission conditions, we enhance the accuracy and consistency of the learned solutions across the entire domain. This approach allows us to effectively address complex problems by decomposing them into smaller, more manageable subdomains, while ensuring the continuity and compatibility of the solutions at the interfaces. Through the utilization of the PECANN framework and the incorporation of interface transmission conditions, we aim to provide a robust and efficient method for distributed learning of PDE problems.

Optimized Schwarz methods have established the benefits of using Robin type interface conditions with optimized parameters as opposed to adopting purely Dirichlet or Neumann type interface conditions. In our proposed approach, we adopt a generalized interface transmission condition using a convex combination of Neumann and Dirichlet conditions. However, one of the aspects of our proposed approach that distinguishes it from optimized Schwarz methods is that, in our method, the parameters of the interface conditions are inferred as part of the PECANN framework and not prescribed as done in optimized Schwarz methods. As we discuss in section \ref{sec:osm},  in optimized Schward methods, the optimal parameters are derived from canonical problems with a simple decomposition by minimizing the convergence rate. These parameters are then used in complex problems. Another distinguishing aspect of our work is that we pursue a non-overlapping decomposition to tackle both Laplace and Helmholtz equations in a unified fashion, whereas in optimized Schwarz methods, separate transmission conditions are used for Laplace and Helmholtz equations \cite{gander2002optimized, gander2007_two_sided_osm, Gander2014}. 

For ease of presentation, we split the domain $\Omega$ into subdomains $\Omega_1$ and $\Omega_2$ sharing the common interface $\Gamma$. We adopt the following absorbing boundary conditions \cite{Engquist1998, smith1998domain} as a generalized Schwarz alternating method. Note that the Robin type interface condition is a convex combination of Dirichlet and Neumann conditions with parameters to be learned. For the first subdomain $\Omega_1$ we have  
\begin{equation}
\begin{aligned}
    -\nabla (u_1 ^{n+1})  &= s_1 &&\text{in} \quad \Omega_1,\\
    u_1^{n+1} &= 0  &&\text{on} \quad \partial \Omega_1 \cap \partial \Omega,\\
    \alpha_1 u_1 ^{n+1} + (1 - \alpha_1) \frac{\partial u_1 ^{n+1}}{\partial \boldsymbol{n_1}}  &= \alpha_1 u_2 ^{n} + (1 - \alpha_1)\frac{\partial u_2 ^{n}}{\partial \boldsymbol{n_1}}  &&\text{on} \quad \Gamma, \label{eq:gsam-1}
\end{aligned}
\end{equation}
and for the second subdomain $\Omega_2$ we have
\begin{equation}
\begin{aligned}
    -\nabla (u_2 ^{n+1})  &= s_2 &&\text{in} \quad \Omega_2,\\
    u_2^{n+1} &= 0  &&\text{on} \quad \partial \Omega_2 \cap \partial \Omega,\\
    \alpha_2 u_2^{n+1} + (1 - \alpha_2)\frac{\partial u_2^{n+1}}{\partial \boldsymbol{n_2}}  &= 
    \alpha_2 u_1^{n} + (1 -\alpha_2) \frac{\partial u_1^{n}}{\partial \boldsymbol{n_2}}  &&\text{on} \quad \Gamma, \label{eq:gsam-2}
\end{aligned}
\end{equation}
where $\alpha_1 > 0 $ and $\alpha_2 > 0 $ are ``learnable'' scalar parameters of the transmission conditions for subdomain $\Omega_1$ and $\Omega_2$ respectively.  We initialize our $\alpha_1^0 = \alpha_1^0 = 1/2$ so as not to favor either the Neumann or the Dirichlet condition. We propose independent parameters (i.e., $\alpha_1$ \& $\alpha_2$) for each subdomain because our solution and its gradient may change significantly across our domains, and having the same $\alpha$ for all the subdomains may not be desirable. Therefore, as the solution improves in each subdomain, $\alpha_1$  and $\alpha_2$ evolve toward independent optimal values. Consequently, setting them as independent parameters enables us to readily learn these parameters for any complex problem. Equally important, through this strategy, each subdomain can exchange information across its interface while minimizing its mismatch with its neighboring subdomain. We should mention that using different parameters in transmission conditions sharing the same interface is not uncommon. For instance, \citet{gander2007_two_sided_osm} used different parameters in the Robin transmission conditions on subdomains sharing a common interface and showed that the resulting domain decomposition method with different parameters performs better than using the same parameters in the Robin transmission conditions.

Next, we present our PECANN formulation with domain decomposition using the generalized Schwarz alternating method given by Eq. \ref{eq:gsam-1} - \ref{eq:gsam-2}. For ease presentation, we split the spatial domain into two subdomains, but our method can handle multiple subdomains. In the following equations, $\mathcal{J}_i(\theta_i)$ is the objective function representing the governing partial differential equation in domain $\Omega_i$, $\mathcal{C}_{1}(\theta_i)$, $\mathcal{C}_{2}(\theta_i)$ are the expected equality constraint functions due to physical boundary conditions and interface transmission conditions, respectively. Subscript $i$ is the subdomain index resulting from the partitioning of the domain. For the first subdomain $\Omega_1$ we have
\begin{linenomath}
\begin{align}
\mathcal{J}_1(\theta_1) &:=  \frac{1}{N_{\Omega_1}} \sum_{i=1}^{N_{\Omega_1}} \|\nabla (u_1 ^{n+1}) - s_1 \|_2^2  \quad ~\text{in} \quad \Omega_1,\\
\mathcal{C}_1(\theta_1) &:= \frac{1}{N_{\partial \Omega_1}} \sum_{i=1}^{N_{\partial \Omega_1}} \|u_1 ^{n+1} - g_1 \|_2^2 \quad ~ \text{on} \quad \partial \Omega_1 \cap \partial \Omega,\\
 \mathcal{C}_2(\theta_1) &:=\frac{1}{N_{\Gamma}} \sum_{i=1}^{N_{\Gamma}} 
\| \alpha_1 ( u_1 ^{n+1} -u_2 ^{n}) \|_2^2 + \| (1 - \alpha_1) (\frac{\partial u_1 ^{n+1}}{\partial \boldsymbol{n_1}} -\frac{\partial u_2 ^{n}}{\partial \boldsymbol{n_1}} )\|_2^2 \quad ~\text{on} \quad \Gamma,
\end{align}
\end{linenomath}
and similarly for the second subdomain $\Omega_2$ we have
\begin{linenomath}
\begin{align}
\mathcal{J}_2(\theta_2) &:=\frac{1}{N_{\Omega_2}} \sum_{i=1}^{N_{\Omega_2}} \|\nabla (u_2 ^{n+1}) - s_2 \|_2^2  \quad \text{in} \quad \Omega_2,\\
\mathcal{C}_1(\theta_2) &:=\frac{1}{N_{\partial \Omega_2}} \sum_{i=1}^{N_{\partial \Omega_2}} \|u_2 ^{n+1} - g_2 \|_2^2 \quad \text{on} \quad \partial \Omega_2 \cap \partial \Omega,\\
\mathcal{C}_2(\theta_2) &:=\frac{1}{N_{\Gamma}} \sum_{i=1}^{N_{\Gamma}} 
\| \alpha_2 ( u_2 ^{n+1} -u_1 ^{n}) \|_2^2 + \| (1 - \alpha_2) (\frac{\partial u_2 ^{n+1}}{\partial \boldsymbol{n_2}} -\frac{\partial u_1^{n}}{\partial \boldsymbol{n_2}} )\|_2^2  \quad ~ \text{on}~ \quad \Gamma,
\end{align}
\end{linenomath}
\begin{algorithm}[t]
    \SetKwInOut{Input}{Input}
    \SetKwInOut{Output}{Output}
    \Input{Collocation points $D$, number of subdomains $K$, number of epochs $E$, number of outer iterations $T$}
    \For{$k \gets 1$ to $K$}{
    
        Initialize subdomain $k$ and assign it a portion of the global problem\;
        Initialize the local model for subdomain $k$\;
        Initialize the Robin parameter $\alpha_k$\;
        Initialize Lagrange multipliers $\lambda_i$ for each type of constraint function\;
        Initialize penalty parameters $\mu_i$ for each type of constraint function\;
    }
    
    \For{$t \gets 1$ to $T$}{
    
        \For{$k \gets 1$ to $K$}{
        Train local model for subdomain $k$ for $E$ epochs independently
        }
        
        Exchange interface information between neighboring models\;
        Reset Lagrange multipliers for interface constraints;
    }
    \Output{Trained local models}
    \caption{Domain Decomposition Training Procedure} \label{alg:ddm}
\end{algorithm}
The unconstrained objective function (i.e. augmented Lagrangian) for each subdomain is then formed through Eq. \ref{eq:unconst_problem}. 

Algorithm \ref{alg:ddm} is our domain decomposition training procedure for solving PDEs using deep learning. Input to the algorithm are collocation points, the number of subdomains $K$, the number of epochs $E$ for local training, and the number of outer iterations $T$ for DDM.
The algorithm initializes each subdomain $k$ with a portion of the global problem, a local model, a Robin parameter $\alpha_k$, vector of Lagrange multipliers, and penalty parameters. It then trains each local model in parallel and exchanges interface information between neighboring models. The interface Lagrange multipliers are reset at each outer iteration. 
The output of the algorithm is a set of trained local models. The main idea of the algorithm is to divide the global problem into subdomains and solve each subdomain separately, exchanging information at the end of each local training. This approach allows a trade-off between communication and computation, making it suitable for distributed computing environments.

\section{Application to Forward PDE problems}
Poisson's and Helmholtz equations have key significance in the field of domain decomposition methods. Discretization of Poisson's equation with a suitable numerical scheme creates a symmetric positive definite matrix whereas, in the case of a Helmholtz equation, which governs propagation phenomena, the resulting matrix is symmetric but non-positive \cite{dolean2015introduction}. Furthermore, it has been established that classical Schwarz method works for Poisson's equation only when there is overlap of subdomains and the convergence of the method depends on the width of the overlap. Whereas for the Helmholtz equation, the classical Schwarz does not converge, even with overlap \cite{gander2006optimized}. Therefore, separate transmission conditions have been proposed to solve Poisson's and Helmholtz equations with domain decomposition.

In the following examples, we apply our proposed DDM to both the Poisson's and Helmholtz equations without any modification to demonstrate the effectiveness of our approach for physics-constrained machine learning of PDEs.
\begin{figure}
    \centering
    \subfloat[]{\includegraphics[scale=0.97]{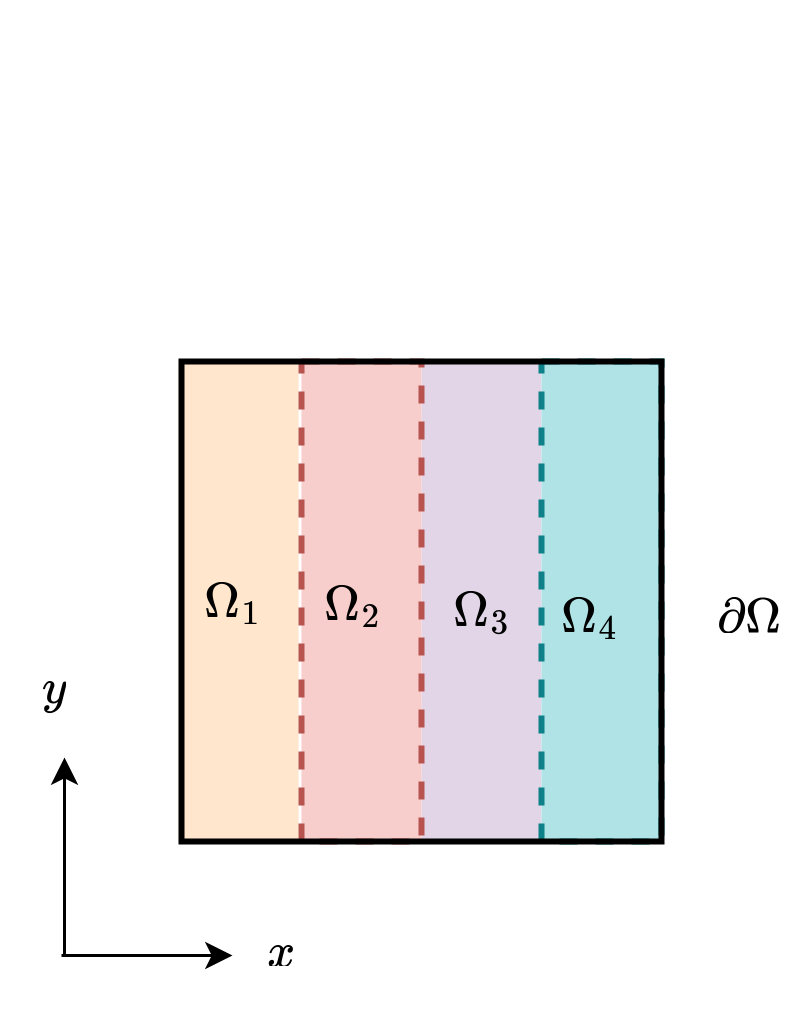}} \qquad
\subfloat[]{\includegraphics{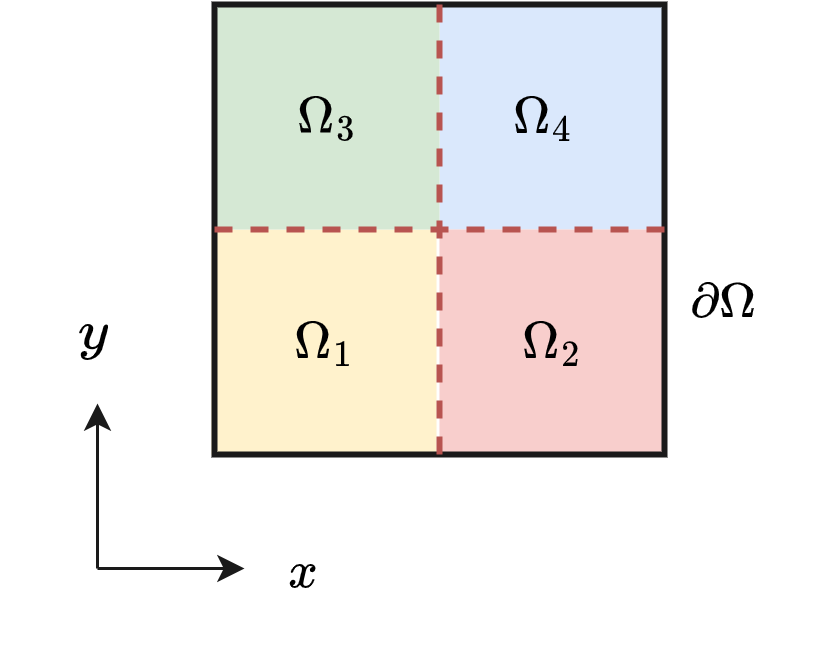}}
    \caption[Domain splitting types]{Domain splitting types: (a) one-way splitting with non-overlapping subdomains, (b) two-way splitting with non-overlapping subdomains}
    \label{fig:domains}
\end{figure}

\subsection{Poisson's Equation}
We consider the following Poisson's equation on the domain $\Omega = \{ (x,y) ~ | ~ -1 \le x \le 1, -1 \le y \le 1 \}$  
\begin{subequations}
\begin{align}
        \nabla^2 u &= s, ~ \text{in} ~ \Omega,\\
         u &= g, ~ \text{on} ~ \partial \Omega,
\end{align}
 \label{eq:poisson_2d}
\end{subequations}
where $\nabla^2$ is the Laplacian operator applied to the function $u$, and $s$ is a given source term, and $\partial \Omega$ is the boundary of the domain $\Omega$. 

We manufacture an oscillatory solution that satisfies Eq.~\eqref{eq:poisson_2d} as follows:
\begin{equation}
u(x,y) = \sin( \frac{\pi}{2} x - \frac{\pi}{2}) \sin( \frac{\pi}{2} y - \frac{\pi}{2}), \quad \forall (x,y) \in \Omega.
 \label{eq:poisson_2d_exact}
\end{equation}
 The corresponding source functions $s(x,y)$ and $g(x,y)$ can be calculated exactly by substituting the manufactured solution into Eq.~\eqref{eq:poisson_2d}. 

For this problem, we utilize a feed-forward neural network consisting of three hidden layers, with each layer containing 20 neurons for each subdomain. The neural network models are designed to have two inputs and one output and employs the tangent hyperbolic activation function. We train our local neural network models for 500 epochs before exchanging the interface information. It should be emphasized that the Poisson equation is an elliptic PDE which lacks any characteristic curves. Inefficient domain decomposition methods may require excessive number of communications and information exchanges between neighboring subdomains to achieve convergence or satisfactory accuracy. However, this can lead to substantial communication overhead, resulting in increased computational complexity and time requirements. Additionally, excessive communication can undermine the advantages of domain decomposition, as the overall efficiency gains from parallel processing may be negated by frequent information exchanges and synchronization demands. We limit the outer iteration count to 30, which implies that only 30 communications occur between neighboring subdomains during the process of learning the global solution. To generate the necessary collocation points, we randomly select 1024 points from within each subdomain, and an additional 128 points are selected along each boundary or interface edge only once.
 
\begin{figure}
\centering
    \subfloat[]{\includegraphics[scale=0.55]{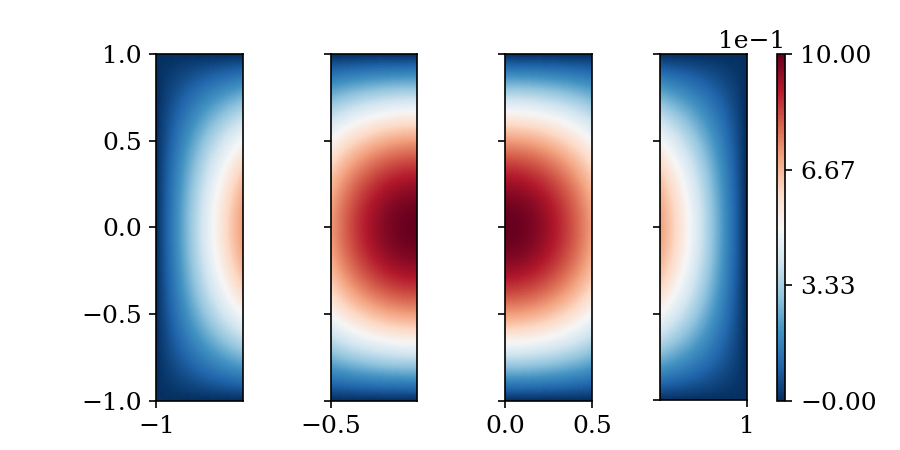}}
    \subfloat[]{\includegraphics[scale=0.55]{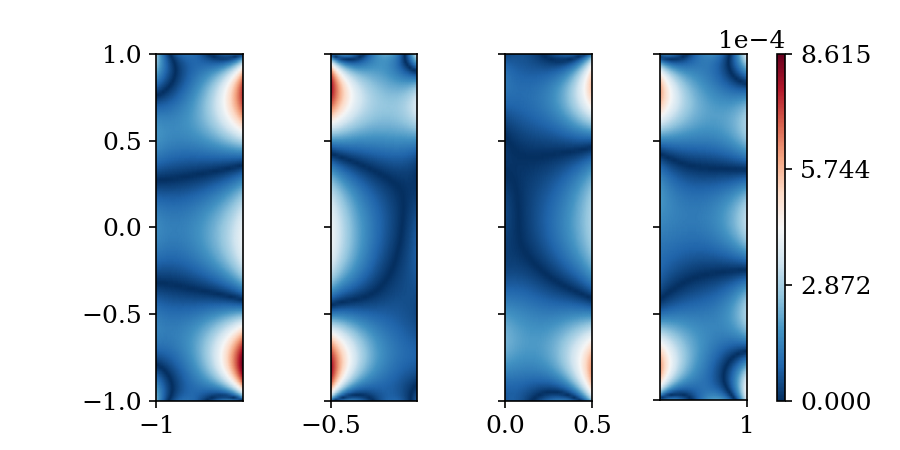}}
    \caption[Poisson's equation on one-dimensional decomposed domain]{Poisson's equation on a one-dimensional decomposed domain: (a) predicted solution on each subdomain, (b) point-wise absolute error on each subdomain}
    \label{fig:poisson_one_way_slicing}
\end{figure}

In our first experiment, we adopt one-dimensional domain decomposition to discover the solution of Poisson's equation. We divide the global domain into four subdomains along one direction, and each subdomain shares a common face with its neighbor. We use the solution at the shared face as an interface condition for the neighboring subdomain. Our aim with this particular decomposition is to demonstrate that boundary conditions propagate across subdomains and middle subdomains that do not have direct access to the physical boundaries are informed by the imposed boundary conditions. This is crucial because it ensures that the solution of Poisson's equation remains accurate and consistent across all subdomains. We present our results in Figure ~\ref{fig:poisson_one_way_slicing}. Our results demonstrate the effectiveness of our method on a one-dimensional domain decomposition for the solution of Poisson's equation while maintaining great accuracy and consistency across subdomains.

\begin{table}
\centering
\caption{Performance comparison of trained models with adaptive and constant Robin penalty parameter $\alpha$}
\label{tb:poisson_one_way}
\vspace{2pt}
\resizebox{0.90\textwidth}{!}{%
\begin{tabular}{@{}lcccc@{}}
\toprule
\multicolumn{1}{c}{Robin Penalty} &
  \multicolumn{1}{c}{maximum $\mathcal{E}_r(u,\hat{u})$ across subdomains} &
  \multicolumn{1}{c}{maximum $\mathcal{E}_\infty(u,\hat{u})$ across subdomains} &
  \\ \midrule
constant $\alpha$ & $1.275 \times 10^{-3}$& $1.430 \times 10^{-3}$ \\
adaptive $\alpha$ & $\boldsymbol{6.245 \times 10^{-4}}$ & $\boldsymbol{7.451 \times 10^{-4}}$ \\
\end{tabular}}
\end{table}
Table \ref{tb:poisson_one_way} shows a performance comparison of trained models with adaptive and constant Robin penalty parameter $\alpha$, in terms of the maximum error across subdomains for two different error measures: $\mathcal{E}r(u,\hat{u})$ and $\mathcal{E}_{\infty}(u,\hat{u})$. The results indicate that the adaptive penalty parameter outperforms the constant penalty parameter, with a significant reduction in maximum error across subdomains for both error measures. The final learned Robin parameters for the subdomains are: $\alpha_1 = 0.6699$, $\alpha_2 = 0.4524$, $\alpha_3 = 0.4564$, and $\alpha_4 = 0.6470$. Notably, $\alpha_1 > 0.5$, indicating a focus on exchanging Dirichlet Conditions, while $\alpha_2 < 0.5$, suggesting an emphasis on matching flux. This novel insight reveals that neighboring subdomains exhibit different evolutions in their Robin parameter behaviors, with varying tendencies towards Dirichlet or Neumann conditions.

\begin{figure}
\centering
    \subfloat[]{\includegraphics[scale=0.5]{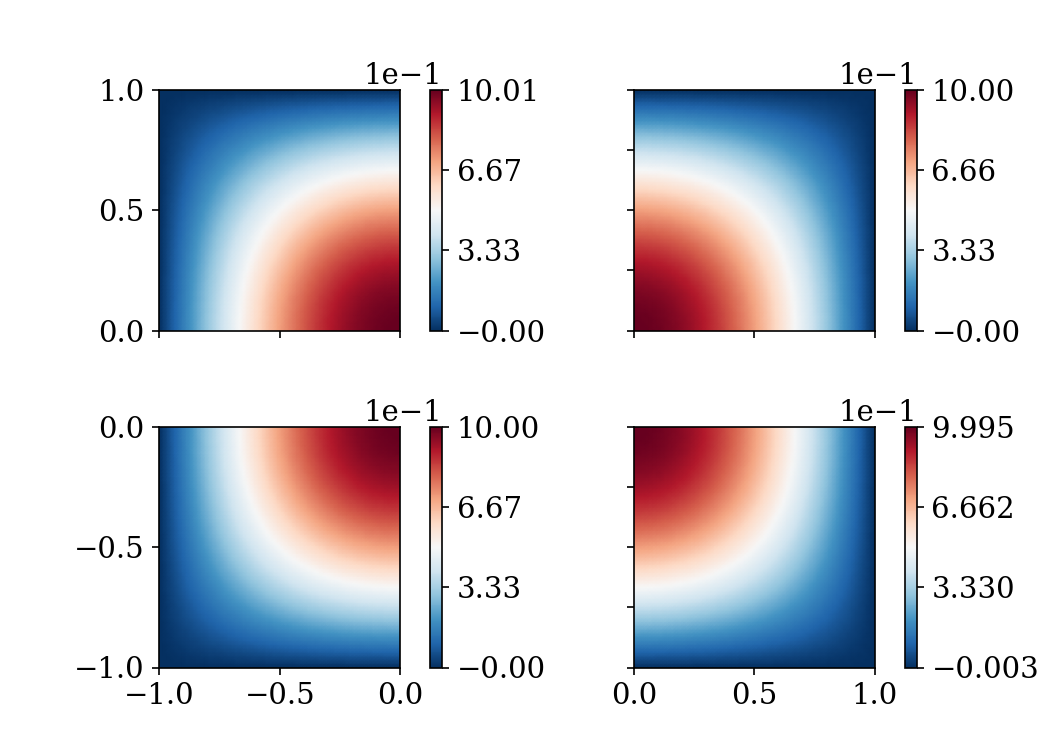}}
    \subfloat[]{\includegraphics[scale=0.5]{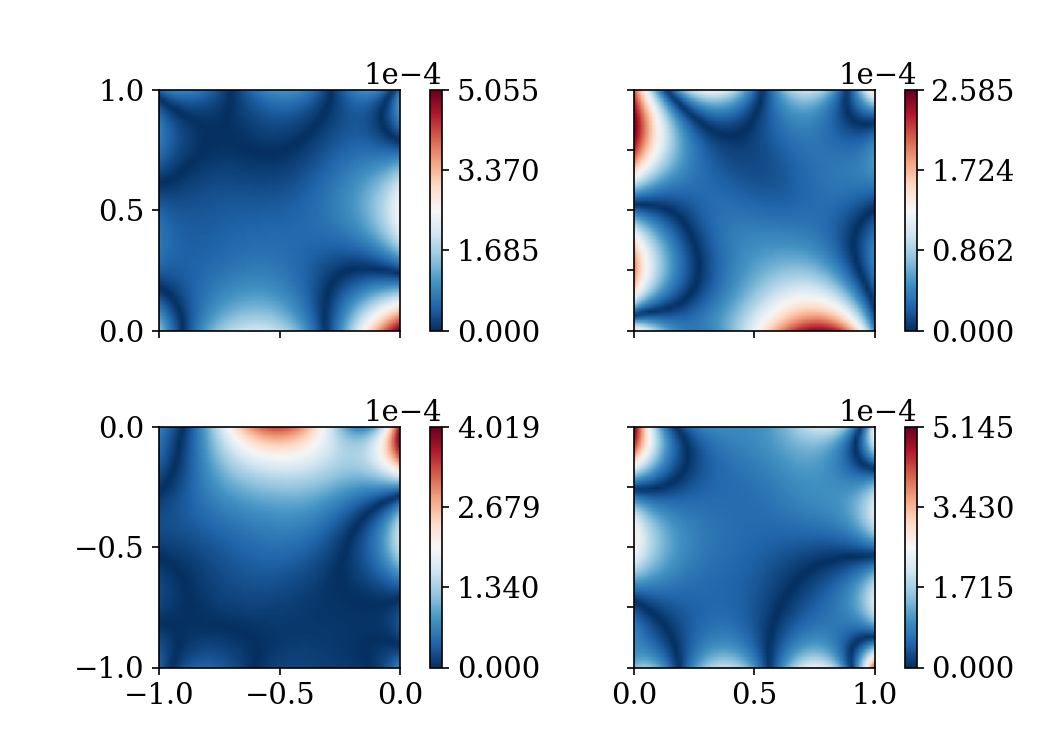}}
    \caption[Solution of Poisson's equation on a two-dimensional Cartesian decomposed domain]{Poisson's equation on two-dimensional Cartesian decomposed domain: (a) exact solution on each subdomain, (b) point-wise absolute error.}
    \label{fig:poisson_two_way_slicing}
\end{figure}

To further investigate the effectiveness of our domain decomposition method, we consider a two-dimensional Cartesian domain decomposition with four subdomains for the same Poisson's equation as in our first experiment. The primary aim is to create a cross point where subdomains meet and can communicate with each other. At this cross point, each subdomain should communicate with all the connecting subdomains to ensure the accuracy and consistency of the solution. We show that by just exchanging information between neighboring subdomains, we can obtain excellent results with a two-dimensional Cartesian domain decomposition. Specifically, we demonstrate that the boundary conditions propagate correctly in both directions, and the solution remains accurate and consistent across all subdomains as can be seen in Figure ~\ref{fig:poisson_two_way_slicing}.
The final learned Robin parameters for the subdomains are: $\alpha_1 = 0.4997$, $\alpha_2 = 0.5018$, $\alpha_3 = 0.5036$, and $\alpha_4 = 0.4949$. Notably, $\alpha_1 < 0.5$ indicates a focus on exchanging Neumann Conditions, while $\alpha_2 > 0.5$ suggests an emphasis on matching Dirichlet Conditions. This novel insight reveals that neighboring subdomains exhibit different evolutions in their Robin parameter behaviors, with varying tendencies towards Dirichlet or Neumann conditions. The observed differences in the learned parameters are a consequence of the distinct physical boundary conditions and random initialization of local models. Despite the symmetric partitioning, the unique characteristics of each subdomain lead to divergent Robin parameters that optimize information exchange effectively.

\subsection{Poisson's Equation with a Complex Decomposition}
We now consider the solution of Poisson's equation on a complex-shaped domain with a complex partitioning. The primary objective of this experiment is to demonstrate the versatility of our method, specifically its ability to handle complex subdomain partitioning and subdomains that lack direct access to the domain boundary $\partial \Omega$. 
The first subdomain is represented by the region between the boundaries $\partial \Omega$ and $\Gamma$, where
\begin{equation}
  \partial \Omega = {(x,y) | x = \rho(\theta)\cos(\theta), y = \rho(\theta)\sin(\theta)
  }, \rho(\theta) = 2 + \sin(2\theta)\cos(2\theta) ,~ \forall 0 \leq \theta \leq 2\pi
\end{equation}
and the interface between subdomains
\begin{equation}
    \Gamma = {(x,y)|x = \rho(\theta)\cos(\theta),y=\rho(\theta)\sin(\theta),\rho(\theta) = 1 + 0.5\cos(4\theta)\sin(6\theta)},~ \forall 0 \leq \theta \leq 2\pi
\end{equation} 

\begin{figure}
\centering
    \subfloat[]{\includegraphics[scale=0.85]{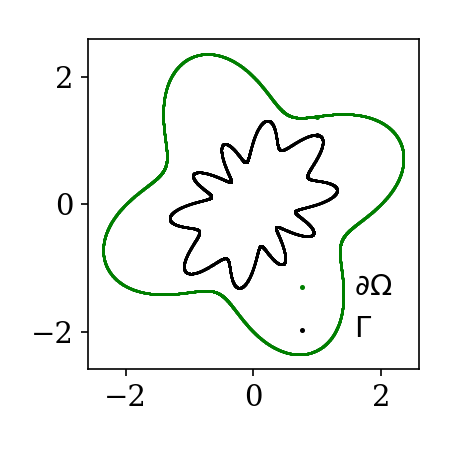}} \hspace{2em}
    \subfloat[]{\includegraphics[trim=0 30 0 0,clip,width=0.5\textwidth]{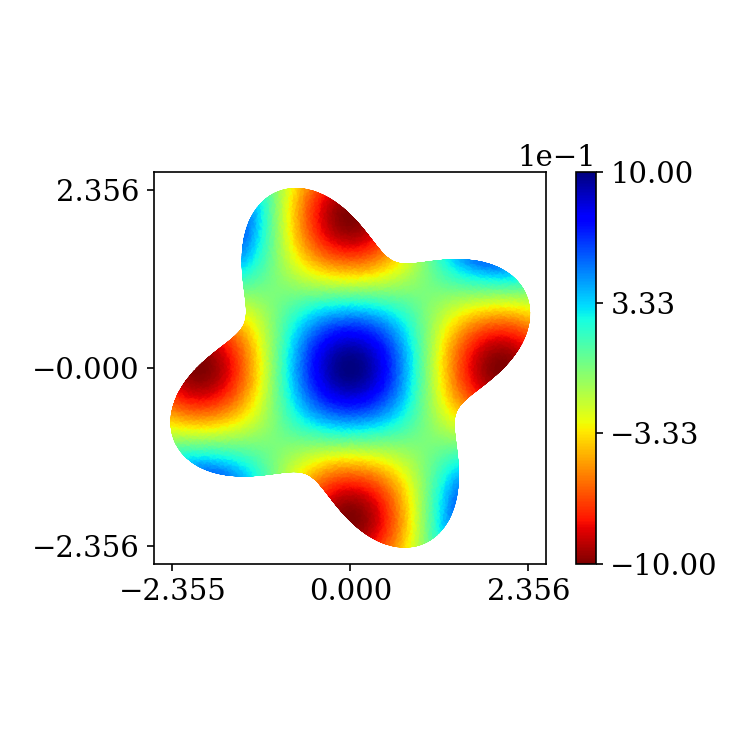}}
       \\
   \subfloat[]{\includegraphics[scale=0.65]{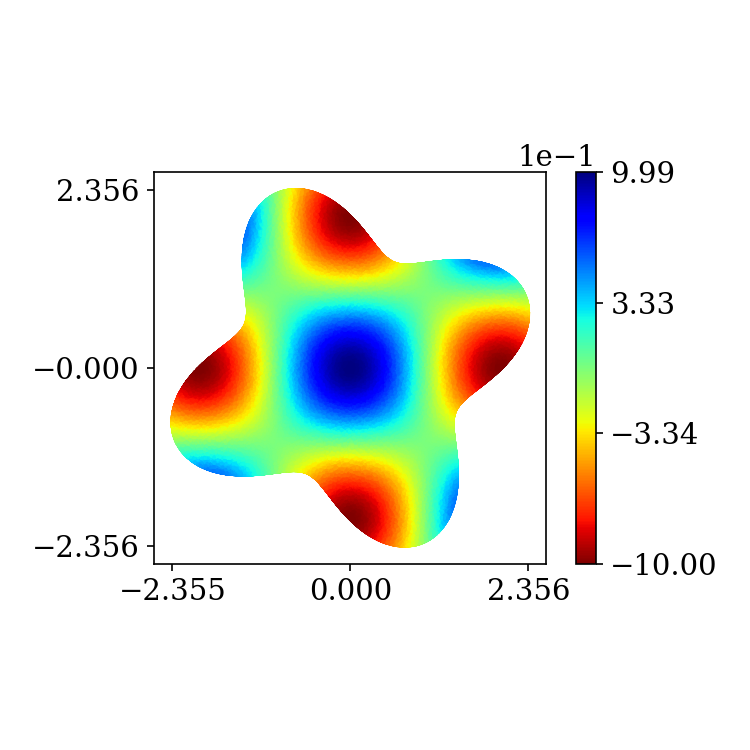}}
    \subfloat[]{\includegraphics[scale=0.65]{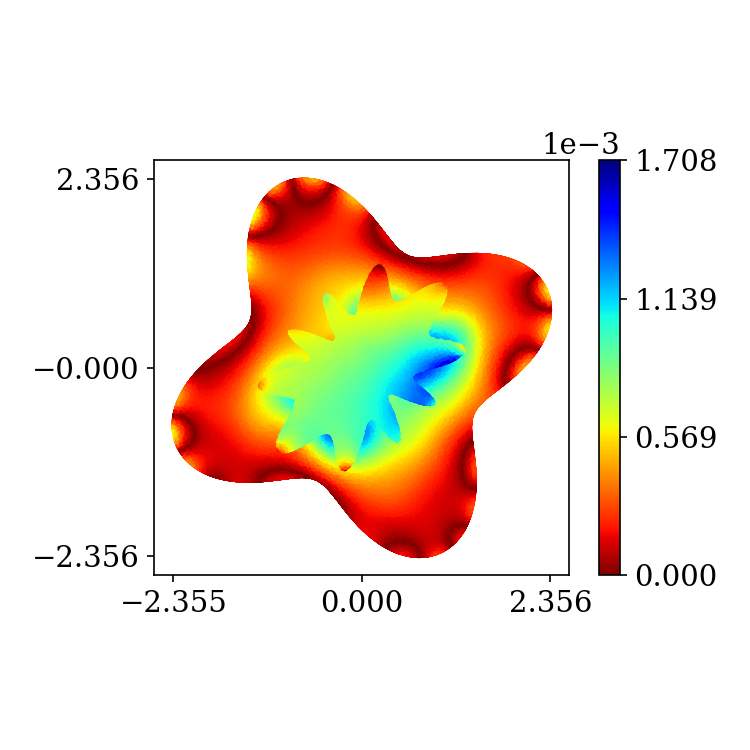}}
    \caption[Steady-state heat conduction in a complex domain and a complex partition]{Steady-state heat conduction in a complex domain with a complex partitioning: (a) complex geometry of domain and the interior subdomain, (b) exact solution, (c) predicted solution on the partitioned domain, (d) point-wise absolute error on each subdomain}
\label{fig:heat_equation_complex_domains}
\end{figure}

The shape of our subdomain is non-trivial, consisting of a region with a complex boundary. The second subdomain is enclosed by the boundary $\Gamma$. Figure~\ref{fig:heat_equation_complex_domains}(a) provides a visual representation of the complex partitioning that we adopt in this problem. To solve the problem at hand, we employ a feed-forward neural network with two hidden layers, each containing 30 neurons for each subdomain. The neural networks are designed to have two inputs and one output and uses the tangent hyperbolic activation function. We locally train the neural network models for 50 epochs, while setting the outer iteration count to $T = 30$. To generate the necessary collocation points, we randomly select 4096 points from within each subdomain, 4096 points along the boundary $\partial \Omega$, and 4096 points along the interface $\Gamma$. This process is performed only once before training, and the same set of collocation points is used throughout the training process. 

Figure~\ref{fig:heat_equation_complex_domains} presents our results for solving Poisson's equation on the two complex subdomains. The exact global solution is displayed in panel (b), while panel (c) presents the predicted solution obtained from our neural network models. Finally, panel (d) shows the absolute error between the exact and predicted solutions. Our results indicate that our approach provides an efficient and accurate means of approximating solutions to Poisson's equation on complex subdomains. Specifically, we observe excellent agreement between the exact and predicted solutions, highlighting the effectiveness of our approach. Additionally, we have obtained learned values of $\alpha = 0.5058$ for the outer domain and $\alpha = 0.4059$ for the inner subdomain, showcasing the adaptive nature of our neural network model in optimizing information exchange across its interface. The Robin parameter for the outer subdomain, being larger than 0.5, suggests a focus on exchanging Dirichlet information, while the Robin parameter for the interior domain, being smaller than 0.5, indicates a focus on the flux. This observation demonstrates the neural network's ability to tailor the efficiency of information exchange based on the specific requirements of each subdomain, providing valuable insights into the domain's behavior and dynamics. In summary, our approach provides a promising strategy for solving Poisson's equation on complex subdomains without an overlap.

\subsection{Helmholtz Equation}
As we discussed earlier, classical Schwarz methods fail for solving Helmholtz equation with domain decomposition, even with overlapping of subdomains. Furthermore, transmission conditions that work well for the Laplace's equation does not readily extend to handle the Helmholtz equation. Therefore, solving the Helmholtz equation with domain decomposition methods presents a significant challenge. We consider the following Helmholtz equation on the domain $\Omega = \{ (x,y) ~ | ~ -1 < (x,y) < 1 \}$ 
\begin{linenomath}
\begin{align}
    \nabla^2 u + k^2 u &= s, \quad ~\text{in} ~ \Omega,\\
    u &= g, \quad ~ \text{on} ~ \partial \Omega,
    \label{eq:helmholtz_2d}
\end{align}
\end{linenomath}
where $\nabla^2$ is the Laplacian operator applied to the function $u$, $k$ is the wavenumber, and $s$ is a given source term and $\partial \Omega$ is the boundary of the domain $\Omega$. The function $u$ represents the amplitude of the wave, and the equation is typically solved subject to appropriate boundary conditions.

Following the equation presented above, we manufacture an oscillatory solution that satisfies Eq.~\eqref{eq:helmholtz_2d} as follows:
\begin{equation}
u(x,y) = \sin( \pi x ) \cos( \pi y/2), \forall (x,y) \quad ~\text{in} ~ \Omega.
 \label{eq:helmholtz_exact}
\end{equation}
where  and $\partial \Omega$ is its boundary.

We employ a feed-forward neural network consisting of three hidden layers, with each layer containing 20 neurons for each subdomain. The networks have two inputs and one output and employs the tangent hyperbolic activation function. We train the neural network models locally for 500 epochs while setting the outer iteration count to 30. We generate the necessary collocation points by randomly selecting 1024 points from within each subdomain and an additional 128 points along each boundary or interface edge. This process is performed only once before training. We first illustrate the effectiveness of one-dimensional domain decomposition as in \ref{fig:domains}. 

\begin{figure}
\centering
    \subfloat[]{\includegraphics[scale=0.5]{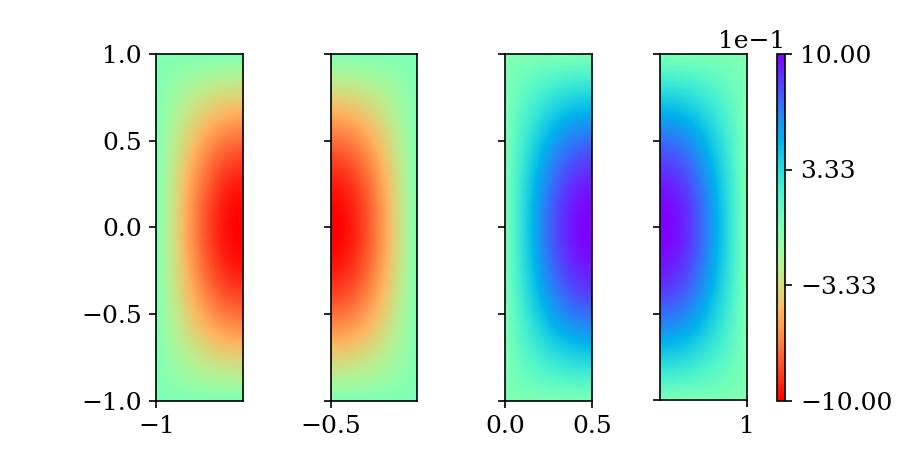}}\quad
    \subfloat[]{\includegraphics[scale=0.5]{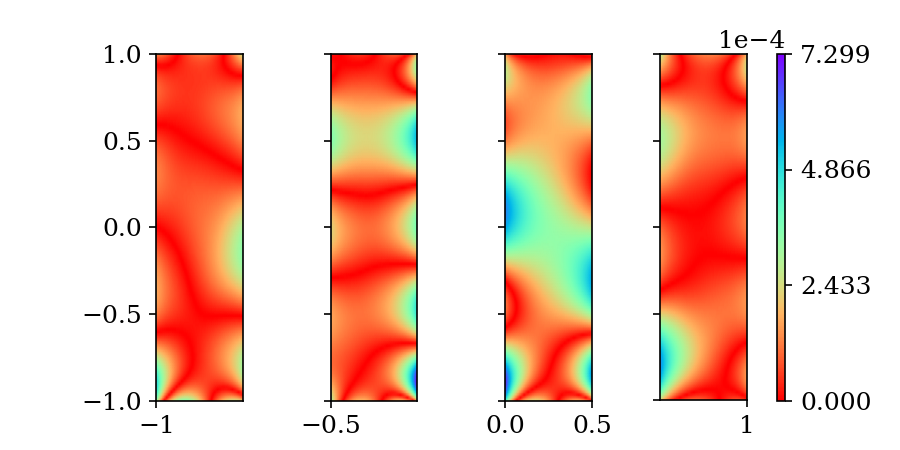}}
    \caption[Helmholtz equation on one-dimensional decomposed domain]{Helmholtz equation on a one-dimensional decomposed domain: (a) predicted solution on each subdomain, (b) absolute point-wise error.}
    \label{fig:helmholtz_one_way_slicing}
\end{figure}

Figure \ref{fig:helmholtz_one_way_slicing} illustrates the results of solving the Helmholtz equation on a one-dimensional decomposed domain. Panel (a) shows the predicted solution obtained from the feed-forward neural network models, while panel (b) shows the absolute error between the exact and predicted solutions. The figure demonstrates the effectiveness of the approach for approximating solutions to the Helmholtz equation on a one-dimensional decomposed domain. 
The final learned Robin parameters for the subdomains are: $\alpha_1 = 0.5522$, $\alpha_2 = 0.7132$, $\alpha_3 = 0.7059$, and $\alpha_4 = 0.5439$. It is interesting to observe that the outer subdomains have larger Robin parameters, indicating a higher focus on matching fluxes than the interior subdomains. This suggests that the neural network effectively adapts to the local characteristics of each subdomain, allocating more importance to matching fluxes within the interior regions for better overall performance in solving the Helmholtz equation.

\begin{figure}
\centering
    \subfloat[]{\includegraphics[scale=0.5]{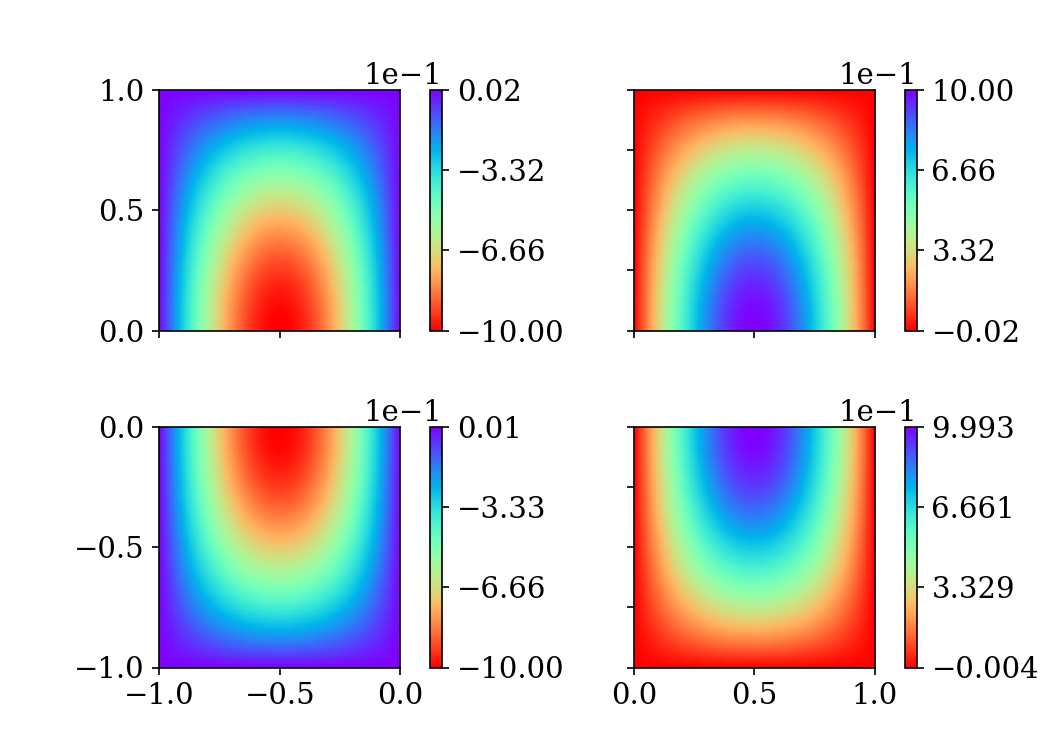}}
    \subfloat[]{\includegraphics[scale=0.5]{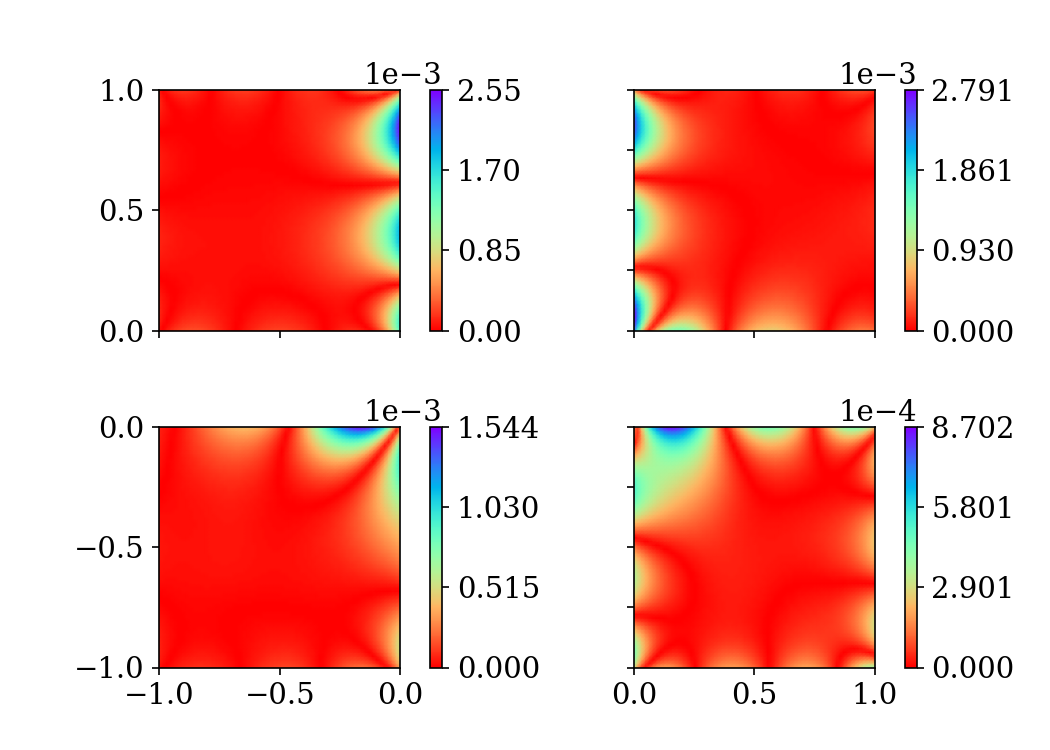}}
    
    \caption[Helmholtz equation on two-dimensional Cartesian decomposed domain]{Helmholtz equation on two-dimensional Cartesian decomposed domain: (a) predicted solution on subdomains, (b) absolute point-wise error}
    \label{fig:helmholtz_two_way_slicing}
\end{figure}

Figure \ref{fig:helmholtz_two_way_slicing} presents the results of solving the Helmholtz equation on a two-dimensional Cartesian decomposed domain using the feed-forward neural network model. Panel (a) displays the predicted solution obtained from local neural network models, while panel (b) shows the absolute error between the exact and predicted solutions. It is evident from the figure that the neural network model can effectively approximate the solutions to the Helmholtz equation on a two-dimensional Cartesian decomposed domain. 
The final learned Robin parameters for the subdomains are: $\alpha_1 = 0.7648$, $\alpha_2 = 0.7577$, $\alpha_3 = 0.7456$, and $\alpha_4 = 0.5521$. Notably, all Robin parameters are larger than 0.5, indicating a focus on exchanging Neumann conditions. This finding aligns with theoretical studies that suggest the incorporation of Neumann conditions is beneficial for solving the Helmholtz equation. The neural network's ability to learn and emphasize the importance of Neumann conditions showcases its adaptability and capability to exploit valuable information for improved accuracy and efficiency in solving the problem.

\section{Application to Inverse Problems}
One of the attractive features of physics-informed/constrained neural networks is that they excel at data-driven and inverse modeling problems.
In an inverse problem, one seeks to determine the unknown parameters or properties of a physical system, such as the conductivity of a material or the distribution of a scalar field, given measurements of some quantity. Inverse problems arise in many fields of engineering. 
In this section, we showcase the versatility and effectiveness of our proposed Domain Decomposition Method (DDM) by applying it to solve inverse problems, akin to the forward problems, without any modifications.

Poisson's equation is expressed 
\begin{linenomath}
\begin{subequations}
\begin{align}
        \nabla^2 u &= s, \quad ~ \text{on} ~ \Omega,\\
         u &= g, \quad ~\text{in} ~ \partial \Omega,
\end{align}
 \label{eq:inverse_poisson_2d}
\end{subequations}
\end{linenomath}
where $\nabla^2$ is the Laplacian operator applied to the function $u$, and $s$ is a given source term. For the inverse Poisson's equation, we use a feed-forward neural network with three hidden layers, each containing 20 neurons for each subdomain. The network has two inputs and one output and employs the tangent hyperbolic activation function. We train the neural network models locally for 500 epochs while setting the outer iteration count to 30. To generate the necessary collocation points, we randomly select 1024 points from within each subdomain, and an additional 128 points are selected along each boundary or interface edge. This process is performed only once before training.

In the context of inverse problems, we consider two different cases. The first case (Case 1) involves a two-dimensional Cartesian subdomain where one of the subdomains lacks physical boundary conditions but has measurement data available. In this case, we aim to demonstrate that we can learn the global solution in the subdomain without having the information on the physical boundary conditions and use the information at the interfaces as interface conditions. 

\begin{figure}
\centering
    \subfloat[]{\includegraphics[scale=0.5]{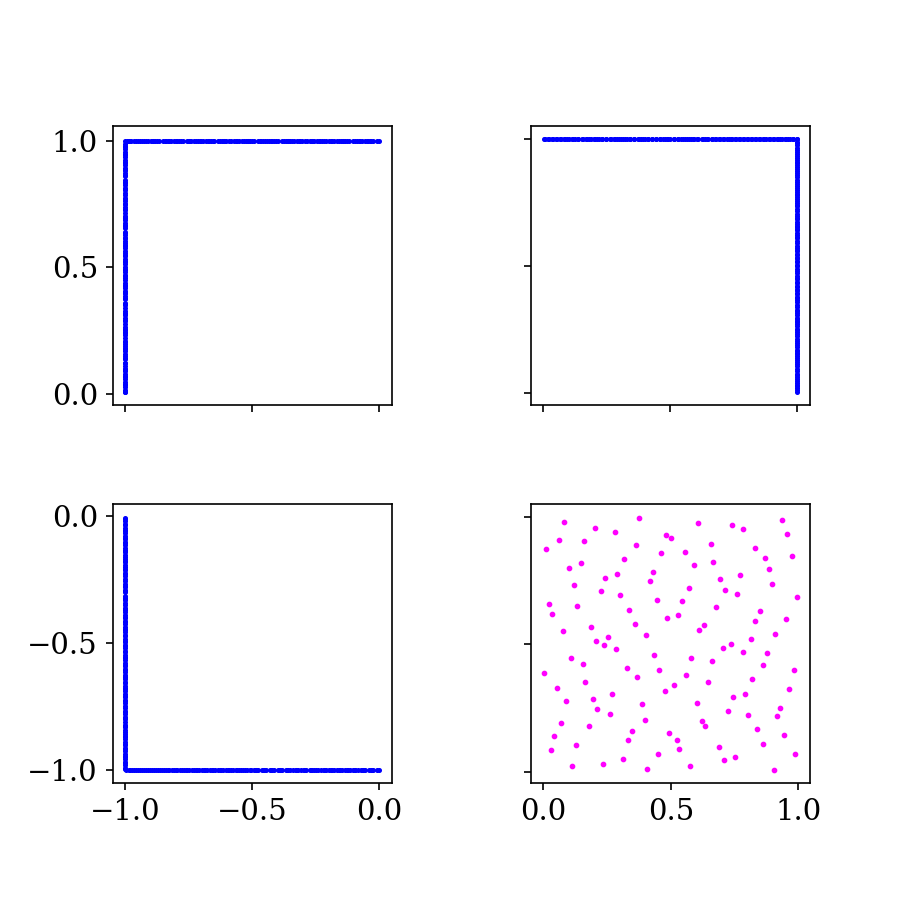}}\\
    \subfloat[]{\includegraphics[scale=0.5]{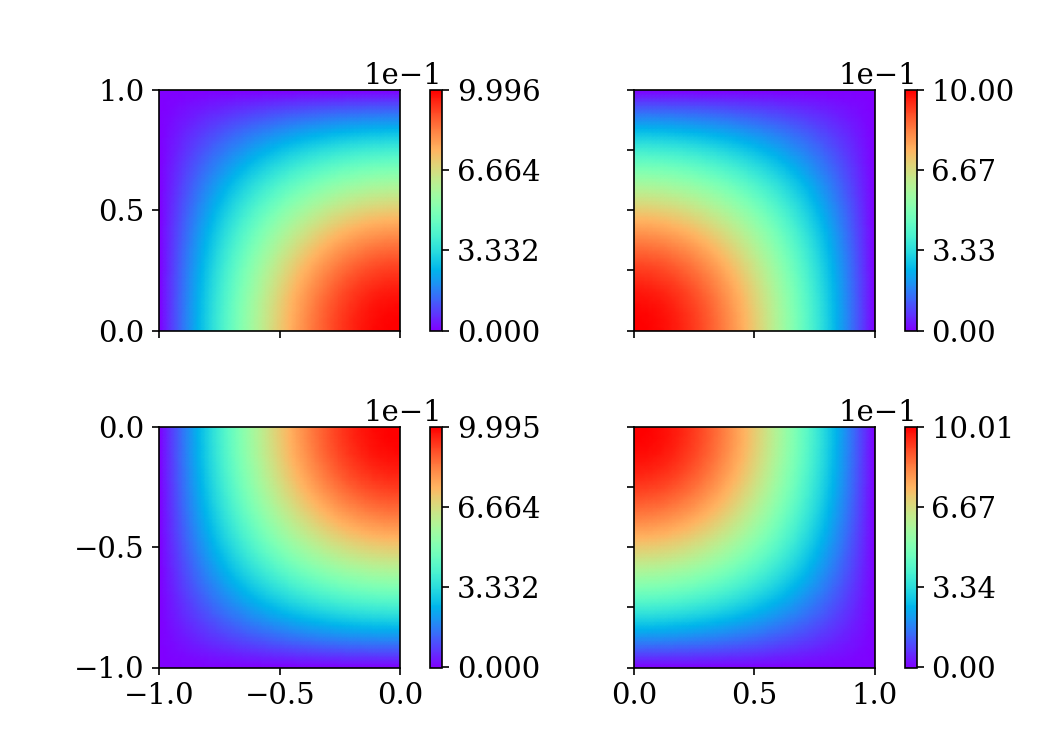}}
    \subfloat[]{\includegraphics[scale=0.5]{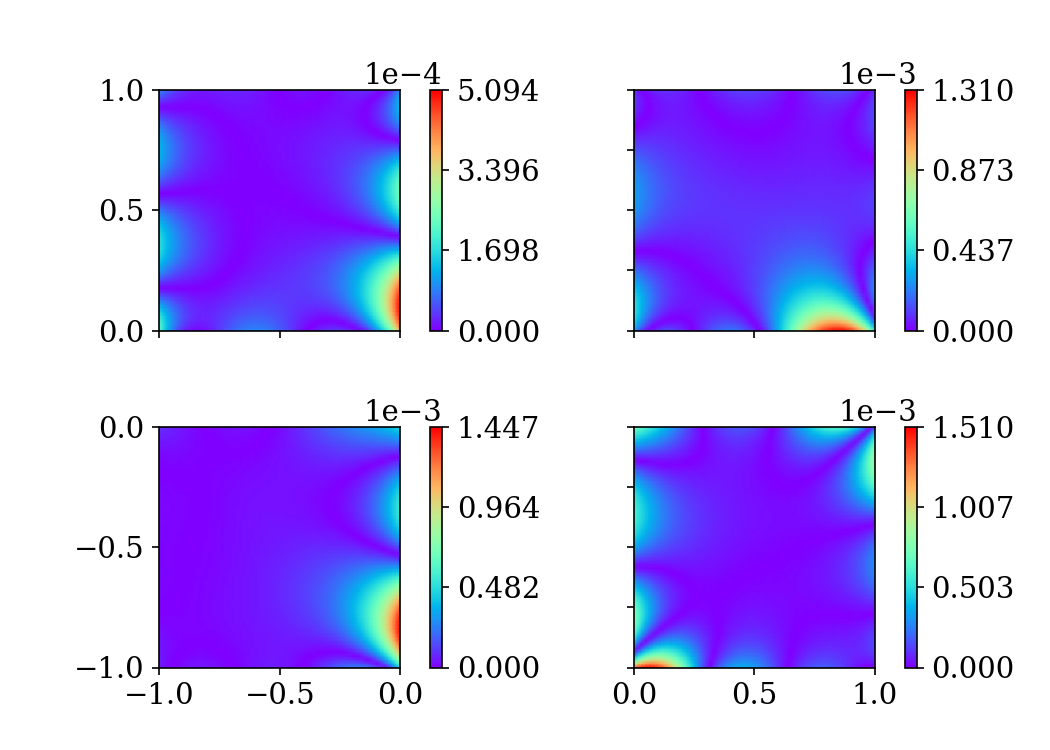}}
    \caption[Inverse Poisson's equation case one]{Inverse Poisson's equation case one: (a) boundary data (blue) and synthetic measurement data (magenta), (b) predicted solution on subdomains, (c) absolute point-wise error}
    \label{fig:inverse_poisson_one}
\end{figure}

\begin{figure}
\centering
    \subfloat[]{\includegraphics[scale=0.5]{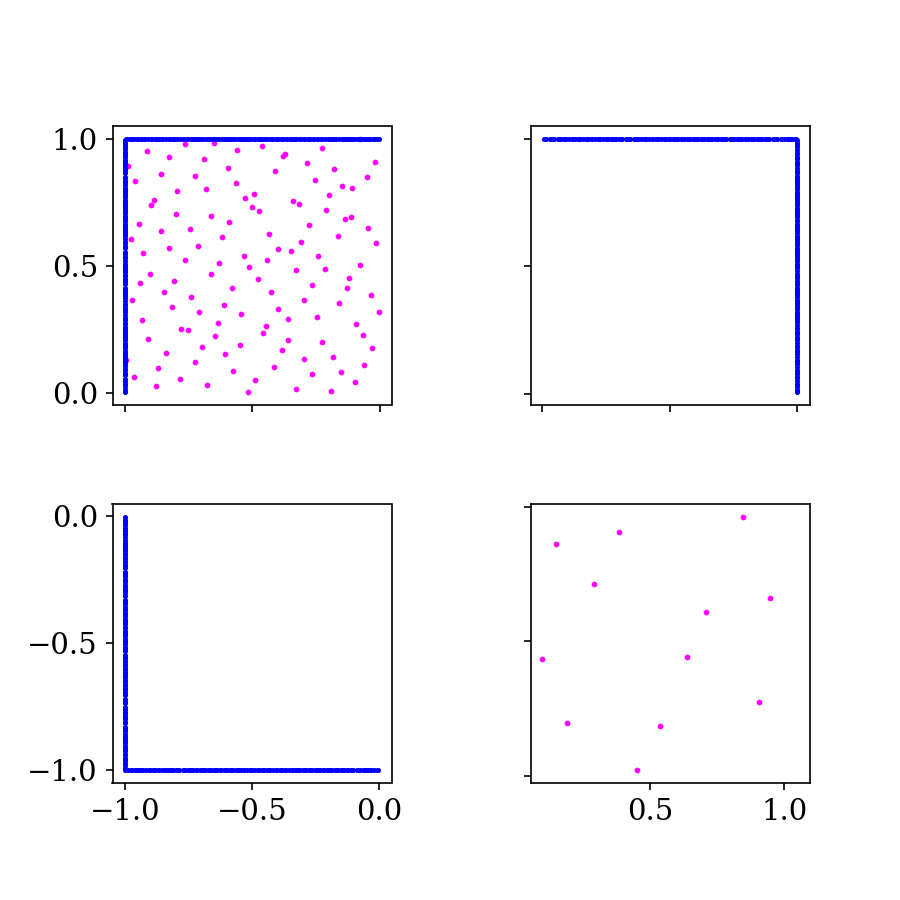}}\\
    \subfloat[]{\includegraphics[scale=0.5]{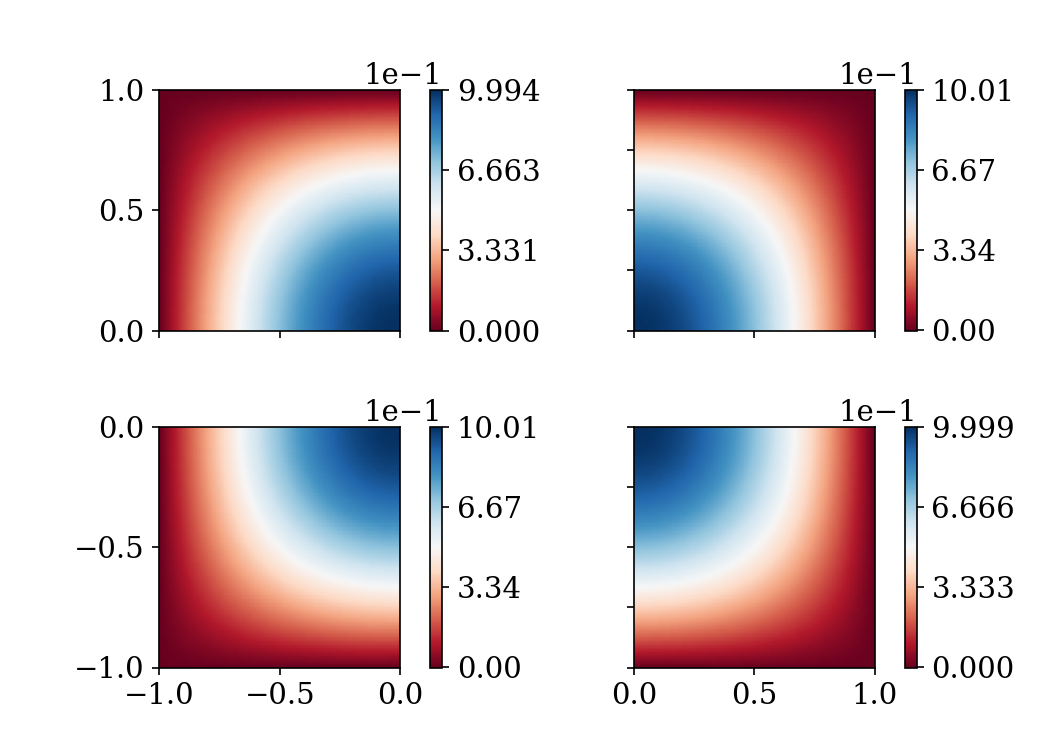}}
    \subfloat[]{\includegraphics[scale=0.5]{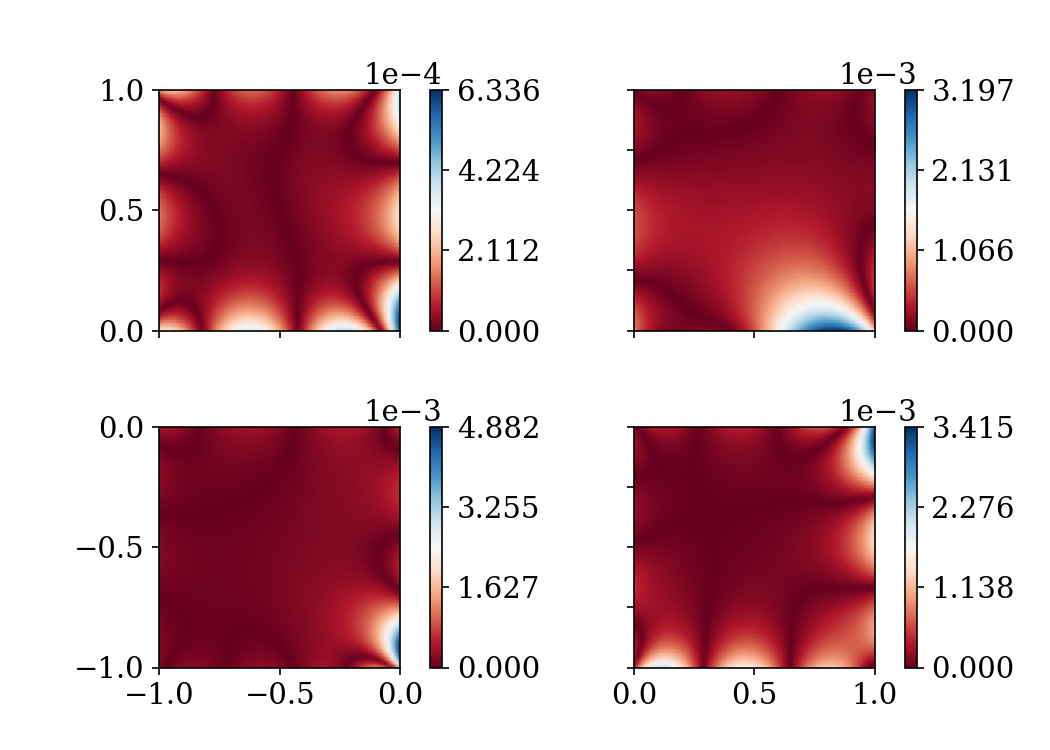}}
    \caption[Inverse Poisson's equation case two]{Inverse Poisson's equation case two: (a) boundary data (blue) and synthetic measurement data (magenta), (b) predicted solution on subdomains, (c) absolute point-wise error}
    \label{fig:inverse_poisson_two}
\end{figure}

Figure \ref{fig:inverse_poisson_one} presents the results of the first case of the inverse problem governed by Poisson's equation. Panel (a) shows the available measurement data as black dots and the subdomains with known boundary conditions in red, while the right bottom subdomain lacks a boundary condition. Panel (b) shows the predicted solution obtained from local neural network models, and panel (c) shows the absolute error between the predicted and true solutions. The figure demonstrates that our models can accurately predict the solution in the right bottom subdomain using the available measurement data. Thus, this approach can effectively solve inverse problems in cases where there are missing boundary conditions but measurement data is available in the subdomain. The final learned Robin parameters for the subdomains are: $\alpha_1 = 0.6667$, $\alpha_2 = 0.5090$, $\alpha_3 = 0.5117$, and $\alpha_4 = 0.5395$. It is interesting to observe that $\alpha_3$ which is the top left subdomain have the smallest Robin parameter among the subdomains and its value is roughly 0.5 indicating an equal focus on matching fluxes and Dirichlet conditions.

In the second case (Case 2) we want to explore how to reconstruct the global solution using only limited measurement data within the subdomain, while the majority of available information is not directly accessible by the subdomain. Figure \ref{fig:inverse_poisson_two} illustrates the results of solving the inverse Poisson's equation in case two. Panel (a) shows the distribution of the labeled data, where the red dots represent the measurement points and the blue dots represent the collocation points. Panel (b) shows the predicted solution and panel (c) shows the absolute error between the predicted and exact solutions. The figure demonstrates that the trained neural network models can accurately discover the solution within the subdomain despite the limited number of available measurements. The final learned Robin parameters for the subdomains are: $\alpha_1 = 0.8173$, $\alpha_2 = 0.5083$, $\alpha_3 = 0.5084$, and $\alpha_4 = 0.5291$. Notably, it is intriguing to observe that $\alpha_1$, which corresponds to the bottom left subdomain, possesses the largest Robin parameter among all the subdomains. This remarkable result suggests a distinct focus on matching Dirichlet conditions in that specific subdomain. Overall, these optimized Robin parameters signify the effectiveness of the learning process in capturing the behavior of the subdomains, and the prominence of $\alpha_1$ emphasizes the significance of Dirichlet boundary conditions in the corresponding region. This improved understanding of the subdomains' characteristics can be valuable for further enhancing the performance and accuracy of the model in relevant applications.

\section{Conclusion}
Domain decomposition methods are needed to extend physics-informed/constrained machine learning methods to solve large-scale problems involving PDEs. In this work, we presented a generalized Schwarz-type domain decomposition method with a Robin-type interface condition to solve forward and inverse PDE problems using physics-informed/constrained neural networks on non-overlapping subdomains. The proposed Robin-type interface condition is a convex combination of Dirichlet and Neumann type interface conditions with a subdomain-specific parameter that we infer or learn as part of the overall solution method. Specifically, we use our previously developed physics and equality constrained artificial neural networks (PECANN) framework \cite{PECANN_2022, basir2023adaptive} to formulate a constrained optimization problem for every local subdomain in which the boundary and subdomain interface conditions act as an equality constraint to the PDE solution within the subdomain. The local constrained optimization formulation is then recast as a dual unconstrained optimization problem using an adaptive augmented Lagrangian method. In our approach, we train a neural network model for each subdomain independently while exchanging information between subdomains through the Robin-type interface condition and discovering its subdomain-specific parameter as part of the training. Although, the interface parameter is discovered as part of the optimization procedure in our approach, our proposed DDM differs from the so-called optimized Schwarz methods in which interface parameters are optimized with respect to the convergence rate of the method.

We have demonstrated the performance and versatility of our method on several forward and inverse problems with various domain partitioning strategies, including complex ones. A noteworthy strength of our proposed DDM coupled with our existing PECANN framework is that it can learn the solution of both the Laplace's and Helmholtz equation with the same transmission conditions and without resorting to any ad-hoc tuning strategies in the neural network model. All the codes accompanying the present work are available as open-source software at \url{https://github.com/HiPerSimLab/PECANN/DDM}.

\section{Acknowledgments}
This material is based upon work supported by the National Science Foundation under Grant No. 1953204 and in part by the University of Pittsburgh Center for Research Computing through the resources provided.

\bibliographystyle{elsarticle-num-names} 
\bibliography{citations}
\end{document}